\documentclass{article}
\usepackage{spconf,amsmath,graphicx}
\usepackage{blindtext}
\usepackage{subcaption}
\usepackage{pifont}
\usepackage{comment}

\title{Slicing Aided Hyper Inference and Fine-tuning \\for Small Object Detection}
\name{
Fatih Cagatay Akyon $^{1,2}$,
Sinan Onur Altinuc $^{1,2}$,
Alptekin Temizel $^{2}$
}
\address{
$^{1}$OBSS AI, Ankara, Turkey\\
$^{2}$Graduate School of Informatics, Middle East Technical University, Ankara, Turkey
}

\usepackage{hyperref, bookmark}
\hypersetup{colorlinks=true,linkcolor=black,citecolor=blue,urlcolor=red}
\begin{document}
%
\maketitle
\begin{abstract}
Detection of small objects and objects far away in the scene is a major challenge in surveillance applications. Such objects are represented by small number of pixels in the image and lack sufficient details, making them difficult to detect using conventional detectors. In this work, an open-source framework called Slicing Aided Hyper Inference (SAHI) is proposed that provides a generic slicing aided inference and fine-tuning pipeline for small object detection. The proposed technique is generic in the sense that it can be applied on top of any available object detector without any fine-tuning. Experimental evaluations, using object detection baselines on the Visdrone and xView aerial object detection datasets show that the proposed inference method can increase object detection AP by 6.8\%, 5.1\% and 5.3\% for FCOS, VFNet and TOOD detectors, respectively. Moreover, the detection accuracy can be further increased with a slicing aided fine-tuning, resulting in a cumulative increase of 12.7\%, 13.4\% and 14.5\% AP in the same order. Proposed technique has been integrated with Detectron2, MMDetection and YOLOv5 models and it is publicly available at \href{https://github.com/obss/sahi.git}{https://github.com/obss/sahi.git}

\end{abstract}
\begin{keywords}
small object detection, sliced inference, windowed inference, visdrone, xview
\end{keywords}

\vspace{-0.3cm}
\section{Introduction}
\vspace{-0.2cm}
\label{sec:intro}

In recent years, object detection has been extensively studied for different applications including face detection, video object detection, video surveillance, self-driving cars. In this field, adoption of deep learning architectures has resulted in highly accurate methods such as Faster R-CNN \cite{ren2015faster}, RetinaNet \cite{lin2017focal}, that are further developed as Cascade R-CNN \cite{cai2018cascade}, VarifocalNet \cite{zhang2021varifocalnet}, and variants. All of these recent detectors are trained and evaluated on well-known datasets such as ImageNet \cite{deng2009imagenet}, Pascal VOC12 \cite{everingham2010pascal}, MS COCO \cite{lin2014microsoft}. These datasets mostly involve low-resolution images ($640\times480$) including considerably large objects with large pixel coverage (covering 60\% of the image height in average). While the trained models have successful detection performances for those types of input data, they yield significantly lower accuracy on small object detection tasks in high-resolution images generated by the high-end drone and surveillance cameras.

\begin{figure}[ht]
\centering
\begin{minipage}[b]{0.15\textwidth}
  \centering
  {\includegraphics[width=1\linewidth]{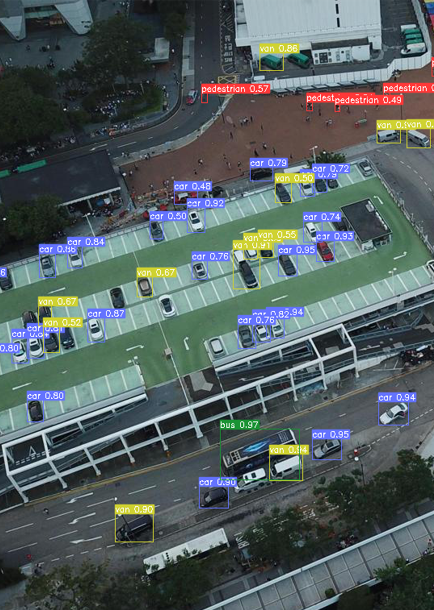}}
\end{minipage}
\begin{minipage}[b]{0.15\textwidth}
  \centering
  {\includegraphics[width=1\linewidth]{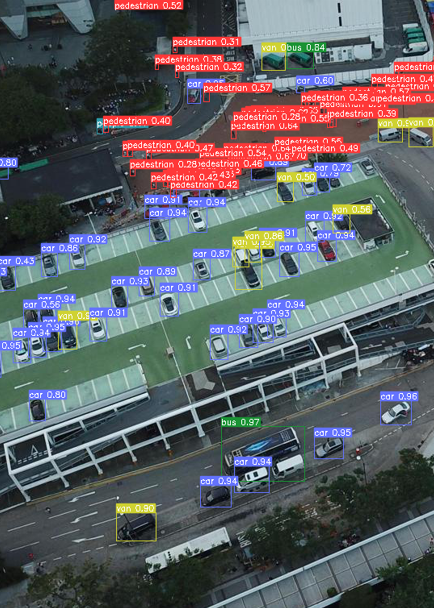}}
\end{minipage}
\begin{minipage}[b]{0.15\textwidth}
  \centering
  {\includegraphics[width=1\linewidth]{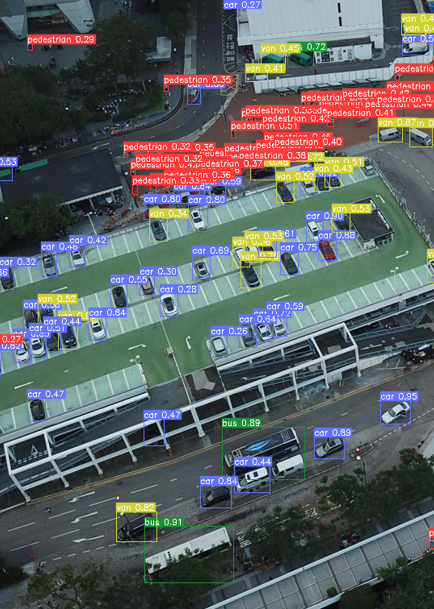}}
\end{minipage}
%
%
\vspace{-0.2cm}
\caption{Results for inference with TOOD detector (left), Slicing-aided hyper inference  (middle), Slicing-aided hyper inference after slicing-aided fine-tuning (right).}
\label{fig:demo}
\vspace{-0.7cm}
\end{figure}

The recent advances in drones, 4K cameras and deep learning research have enabled long-range object detection that is met under Detection, Observation, Recognition and Identification (DORI) criteria \cite{standard2012}. DORI criteria define the minimum pixel height of the objects for different tasks: 10\% of the image height is required to detect and 20\% to recognize the objects (108 pixels in full HD videos). Relatively small pixel coverage pushes the limits of CNN based object detection methods, in addition, high-resolution images demands greater needs in terms of memory requirements.

In this paper, we propose a generic solution based on slicing aided inference and fine-tuning for small object detection on high-resolution images while maintaining higher memory utilization. Fig. \ref{fig:demo} illustrates the improvement of small object detection on a sample image from Visdrone test set. 


\vspace{-0.3cm}
\section{Related Work}
\vspace{-0.2cm}
\label{sec:related}


The recent learning-based object detection techniques can be categorized into two main types. Single-stage detectors, such as SSD \cite{liu2016ssd}, YOLO \cite{bochkovskiy2020yolov4}, RetinaNet \cite{lin2017focal}, directly predict the location of objects without an explicit proposal stage. Two-stage region proposal based methods, such as Fast R-CNN \cite{girshick2015fast}, Faster R-CNN \cite{ren2015faster}, Cascade R-CNN \cite{cai2018cascade}, involve an initial region proposal stage. These proposals are then refined to define the position and size of the object. Typically, single-stage approaches are faster than two-stage, while the latter has higher accuracy.

More recently, anchor-free detectors have started to attract attention. They eliminate the use of anchor boxes and classify each point on the feature pyramid \cite{lin2017feature} as foreground or background, and directly predict the distances from the foreground point to the four sides of the ground-truth bounding box, to produce the detection.
FCOS \cite{tian2019fcos} is the first object detector eliminating the need for predefined set of anchor boxes and entailing computational need.
VarifocalNet (VFNet) \cite{zhang2021varifocalnet} learns to predict the IoU-aware classification score which mixes the object presence confidence and localization accuracy together as the detection score for a bounding box. The learning is supervised by the proposed Varifocal Loss (VFL), based on a new star-shaped bounding box feature representation. 
TOOD \cite{feng2021tood} explicitly aligns the two tasks (object classification and localization) in a learning-based manner utilizing novel task-aligned head which offers a better balance between learning task-interactive and task-specific features and task alignment learning via a designed sample assignment scheme and a task-aligned loss.

The algorithms designed for general object detection perform poorly on high resolution images that contain small and dense objects, leading to specific approaches for small object detection. 
In \cite{wang2019pso}, a particle swarm optimization (PSO) and bacterial foraging optimization (BFO)-based learning strategy (PBLS) is used to optimize the classifier and loss function. However, these heavy modifications to the original models prevent fine-tuning from pretrained weights and require training from scratch. Moreover, due to unusual optimization steps they are hard to adapt into a present detector. The method proposed in \cite{kisantal2019augmentation} oversamples images with small objects and augments them by making several copies of small objects. However, this augmentation requires segmentation annotations and, as such, it is not compatible with the object detection datasets. The method in \cite{chen2019ssd} can learn richer features of small objects from the enlarged areas, which are clipped from the raw image. The extra features positively contribute to the detection performance but the selection of the areas to be enlarged brings a computational burden. In \cite{bosquet2018stdnet}, a fully convolutional network is proposed for small object detection that contains an early visual attention mechanism that is proposed to choose the most promising regions with small objects and their context. In \cite{van2019satellite}, a slicing based technique is proposed but the implementation is not generic and only applicable to specific object detectors. In \cite{pang2019jcs}, a novel network (called JCS-Net) is proposed for small-scale pedestrian detection, which integrates the classification task and the super-resolution task in a unified framework. \cite{bai2018finding} proposed an algorithm to directly generate a clear high-resolution face from a blurry small one by adopting a generative adversarial network (GAN). However, since these techniques propose new detector architectures they require pretraining from scratch with large datasets which is costly.

In contrast to the mentioned techniques, we propose a generic slicing aided fine-tuning and inference pipeline that can be utilized on top of any existing object detector. This way, small object detection performance of any currently available objects detector can be boosted without any fine-tuning (by slicing aided inference). Moreover, additional performance boost can be gained by fine-tuning the pretrained models. 

\vspace{-0.3cm}
\section{Proposed Approach}
\vspace{-0.2cm}
\label{sec:proposed}

In order to handle the small object detection problem, we propose a generic framework based on slicing in the fine-tuning and inference stages. Dividing the input images into overlapping patches results in relative larger pixel areas for small objects with respect to the images fed into the network.  
\begin{figure}[ht]
\centering
\begin{minipage}[b]{0.45\textwidth}
  \centering
  {\includegraphics[width=1.0\linewidth]{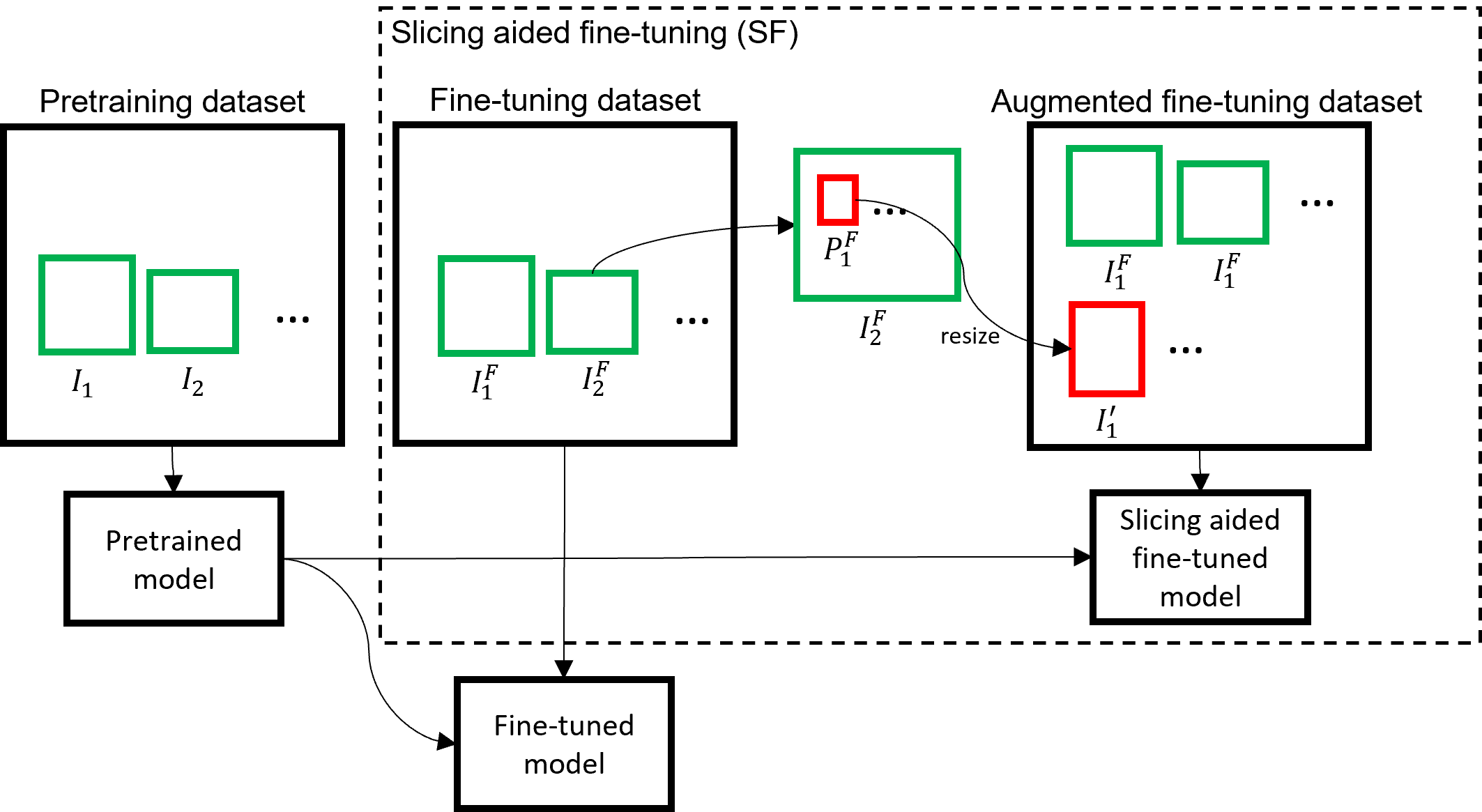}}
\end{minipage}

\hfill

\begin{minipage}[b]{0.45\textwidth}
  \centering
  {\includegraphics[width=1.0\linewidth]{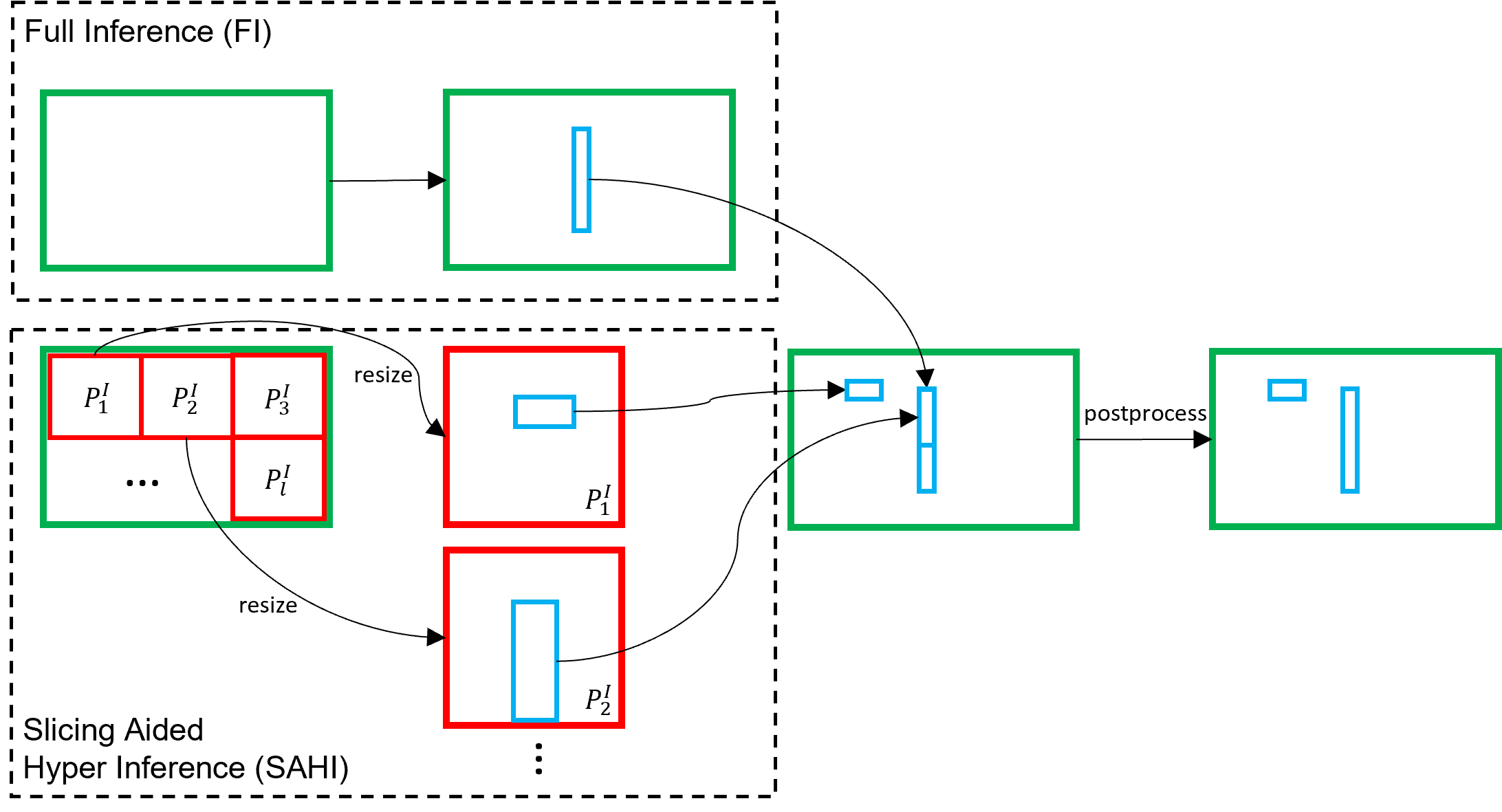}}
\end{minipage}
\vspace{-0.1cm}
\caption{Slicing aided fine-tuning (top) and slicing aided hyper inference (bottom) methods. In finetuning, the dataset is augmented by extracting patches from the images and resizing them to a larger size. During inference, image is divided into smaller patches and predictions are generated from larger resized versions of these patches. Then these predictions are converted back into original image coordinates after NMS. Optionally, predictions from full inference can also be added.}
\label{fig:diagrams}
\vspace{-0.3cm}
\end{figure}

\textbf{Slicing Aided Fine-tuning (SF)}: Widely used object detection frameworks such as Detectron2 \cite{wu2019detectron2}, MMDetection \cite{chen2019mmdetection} and YOLOv5 \cite{jocher2021ultralytics} provide pretrained weights on the datasets such as ImageNet \cite{deng2009imagenet} and MS COCO \cite{lin2014microsoft}. This allows us to fine-tune the model using smaller datasets and over shorter training spans in contrast to training from scratch with large datasets. These common datasets mostly involve low-resolution images $(640\times480)$ having considerably large objects with large pixel coverage (covering 60\% of the image height in average). The models pretrained using these datasets provide very successful detection performance for similar inputs. On the other hand, they yield significantly lower accuracy on small object detection tasks in high-resolution images generated by the high-end drone and surveillance cameras.

In order to overcome this issue, we augment the dataset with by extracting patches from the images fine-tuning dataset as seen in Fig. \ref{fig:diagrams}. Each image $I^F_1, I^F_2,..., I^F_j$ is sliced into overlapping patches $P^F_1, P^F_2, ... P^F_k$ with dimensions $M$ and $N$ are selected within predefined ranges $[M_{min}, M_{max}]$ and $[N_{min}, N_{max}]$ which are treated as hyper-parameters. Then during fine-tuning, patches are resized by preserving the aspect ratio so that image width is between 800 to 1333 pixels to obtain augmentation images $I'_1, I'_2,..., I'_k$, whereby the relative object sizes are larger compared to the original image. These images $I'_1, I'_2,..., I'_k$, together with the original images $I^F_1, I^F_2,..., I^F_j$ (to facilitate detection of large objects), are utilized during fine-tuning. It has to be noted that, as the patch sizes decrease, larger objects may not fit within a slice and the intersecting areas, and this may lead to poor detection performance for larger objects.

\textbf{Slicing Aided Hyper Inference (SAHI):}
Slicing method is also utilized during the inference step as detailed in Fig. \ref{fig:diagrams}. First, the original query image $I$ is sliced into $l$ number of $M\times N$ overlapping patches $P^I_1, P^I_2, ... P^I_l$ . Then, each patch is resized while preserving the aspect ratio. After that, object detection forward pass is applied independently to each overlapping patch. An optional full-inference (FI) using the original image can be applied to detect larger objects. Finally, the overlapping prediction results and, if used, FI results are merged back into to original size using NMS. During NMS, boxes having higher Intersection over Union (IoU) ratios than a predefined matching threshold $T_m$ are matched and for each match, detections having detection probability than lower than $T_d$ are removed.


\vspace{-0.4cm}
\section{Results}
\vspace{-0.2cm}
\label{sec:results}

The proposed method has been integrated into FCOS \cite{tian2019fcos}, VarifocalNet \cite{zhang2021varifocalnet} and TOOD \cite{feng2021tood} object detectors using MMDetection \cite{chen2019mmdetection} framework for experimental evaluation. Related config files, conversion and evaluation scripts, evaluation result files have been publicly provided \footnote{\url{https://github.com/fcakyon/sahi-benchmark}}. All slicing related operations have also been made publicly available to enable integration into other object detection frameworks \footnote{\url{https://github.com/obss/sahi}}. 


VisDrone2019-Detection \cite{du2019visdrone} is an object detection dataset having 8599 images captured by drone platforms at different locations and at different heights. Most of the objects in this dataset are small, densely distributed and partially occluded. There are also illumination and perspective changes in different scenarios. More than 540k bounding boxes of targets are annotated with ten predefined categories: \textit{pedestrian, person, bicycle, car, van, truck, tricycle, awning-tricycle, bus, motor}. Super categories are defined as \textit{pedestrian, motor, car and truck}. The training and validation subsets consists of 6471 and 548 images, respectively which are collected at different locations but in similar environments.

xView \cite{lam2018xview} is one of the largest publicly available datasets for object detection from satellite imagery. It contains images from complex scenes around the world, annotated using bounding boxes. It contains over 1M object instances from 60 different classes. During the experiments, randomly selected 75\% and 25\% splits have been used as the training and validation sets, respectively.

Both of these datasets contain small objects (object width $<1\%$ of image width).

\begin{table}[!h]
    \vspace{-0.1cm}
    \centering
    \small
    \begin{tabular}{ccccc}
    Setup & AP$_{50}$ & AP$_{50}$s & AP$_{50}$m & AP$_{50}$l\\
    \hline
    FCOS+FI & 25.8 & 14.2 & 39.6 & 45.1 \\
    FCOS+SAHI+PO & 29.0 & 18.9 & 41.5 & 46.4 \\
    FCOS+SAHI+FI+PO & 31.0 & 19.8 & 44.6 & 49.0 \\

    FCOS+SF+SAHI+PO & 38.1 & 25.7 & 54.8 & 56.9 \\
    FCOS+SF+SAHI+FI+PO & \textbf{38.5} & \textbf{25.9} & \textbf{55.4} & \textbf{59.8} \\
    \hline
    VFNet+FI & 28.8 & 16.8 & 44.0 & 47.5 \\
    VFNet+SAHI+PO & 32.0 & 21.4 & 45.8 & 45.5 \\
    VFNet+SAHI+FI+PO & 33.9 & 22.4 & 49.1 & 49.4 \\
    VFNet+SF+SAHI+PO & 41.9 & \textbf{29.7} & 58.8 & 60.6 \\
    VFNet+SF+SAHI+FI+PO & \textbf{42.2} & \textbf{29.6} & \textbf{59.2} & \textbf{63.3} \\
    \hline
    TOOD+FI & 29.4 & 18.1 & 44.1 & 50.0 \\
    TOOD+SAHI & 31.9 & 22.6 & 44.0 & 45.2 \\
    TOOD+SAHI+PO & 32.5 & 22.8 & 45.2 & 43.6 \\
    TOOD+SAHI+FI & 34.6 & 23.8 & 48.5 & 53.1 \\
    TOOD+SAHI+FI+PO & 34.7 & 23.8 & 48.9 & 50.3 \\

    TOOD+SF+FI & 36.8 & 24.4 & 53.8 & \textbf{66.4} \\
    TOOD+SF+SAHI & 42.5 & 31.6 & 58.0 & 61.1 \\
    TOOD+SF+SAHI+PO & 43.1 & \textbf{31.7} & 59.0 & 60.2 \\
    TOOD+SF+SAHI+FI & 43.4 & \textbf{31.7} & 59.6 & 65.6 \\
    TOOD+SF+SAHI+FI+PO & \textbf{43.5} & \textbf{31.7} & \textbf{59.8} & 65.4 \\

    \end{tabular}
    \vspace{-0.2cm}
    \caption{Mean average precision values calculated on Visdrone19-Detection test-dev set. SF, SAHI, FI, and PO correspond to slicing aided fine-tuning, slicing aided inference, full image inference, and overlapping patches, respectively.}
    \label{tab:visdrone-result}
\end{table}

\begin{table}[!h]
    \vspace{-0.5cm}
    \centering
    \small
    \begin{tabular}{ccccc}
    Setup & AP$_{50}$ & AP$_{50}$s & AP$_{50}$m & AP$_{50}$l\\
    \hline
    FCOS+FI & 2.20 & 0.10 & 1.80 & 7.30 \\

    FCOS+SF+SAHI & 15.8 & 11.9 & 18.4 & 11.0 \\
    FCOS+SF+SAHI+PO & \textbf{17.1} & \textbf{12.2} & \textbf{20.2} & 12.8 \\
    FCOS+SF+SAHI+FI & 15.7 & 11.9 & 18.4 & 14.3 \\
    FCOS+SF+SAHI+FI+PO & \textbf{17.0} & \textbf{12.2} & \textbf{20.2} & \textbf{15.8} \\
    \hline
    VFNet+FI & 2.10 & 0.50 & 1.80 & 6.80 \\
    
    VFNet+SF+SAHI  & 16.0 & 11.9 & 17.6 & 13.1 \\
    VFNet+SF+SAHI+PO & \textbf{17.7} & \textbf{13.7} & \textbf{19.7} & 15.4 \\
    VFNet+SF+SAHI+FI & 15.8 & 11.9 & 17.5 & 15.2 \\
    VFNet+SF+SAHI+FI+PO & \textbf{17.5} & \textbf{13.7} & \textbf{19.6} & \textbf{17.6} \\
    \hline
    TOOD+FI & 2.10 & 0.10 & 2.00 & 5.20 \\

    TOOD+SF+SAHI  & 19.4 & 14.6 & 22.5 & 14.2 \\
    TOOD+SF+SAHI+PO & \textbf{20.6} & \textbf{14.9} & \textbf{23.6} & 17.0 \\
    TOOD+SF+SAHI+FI & 19.2 & 14.6 & 22.3 & 14.7 \\
    TOOD+SF+SAHI+FI+PO & \textbf{20.4} & \textbf{14.9} & \textbf{23.5} & \textbf{17.6} \\
    \hline
    
    \end{tabular}
    \vspace{-0.2cm}
    \caption{Mean average precision values calculated on xView validation split. SF, SAHI, FI, and PO correspond to slicing aided fine-tuning, slicing aided inference, full image inference, and overlapping patches, respectively.}
    \label{tab:xview-result}
    \vspace{-0.2cm}
\end{table}

\begin{figure}[htb]
\vspace{-0.5cm}
\begin{minipage}[b]{1\linewidth}
  \centering
  \centerline{\includegraphics[width=5.5cm]{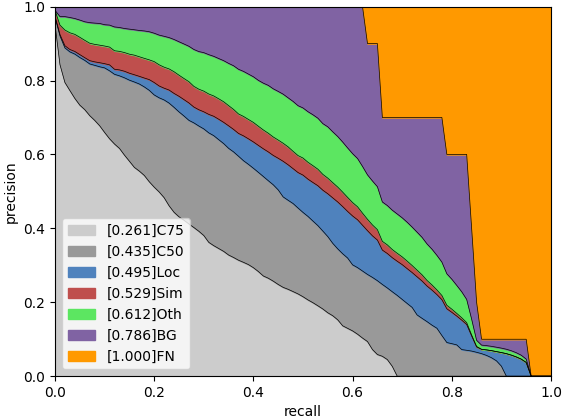}}
\end{minipage}
\vspace{-0.8cm}
\caption{Error analysis curve for TOOD object detector in SF+SAHI setting calculated on Visdrone19-Det test-dev set.}
\label{fig:tood-visdrone-error}
\vspace{-0.3cm}
\end{figure}

\begin{figure}[htb]
\vspace{-0.1cm}
\begin{minipage}[b]{1\linewidth}
  \centering
  \centerline{\includegraphics[width=5.5cm]{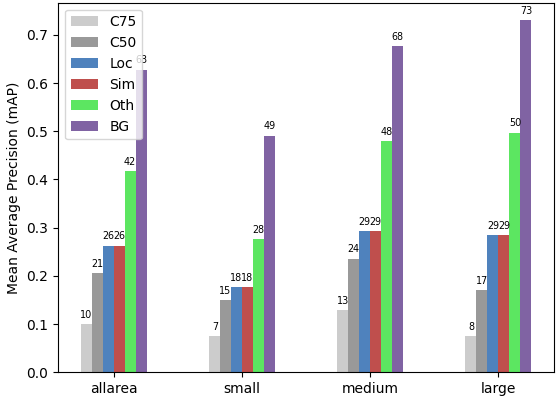}}
  
\end{minipage}
\vspace{-0.7cm}
\caption{Error analysis bar plot for TOOD object detector in SF+SAHI setting calculated on xView validation split.}
\label{fig:tood-xview-error}
\vspace{-0.4cm}
\end{figure}

During experiments, SGD optimizer with a learning rate of $0.01$, momentum of $0.9$, weight decay of $0.0001$ and linear warmup of $500$ iteration is used. Learning rate scheduling is done with exponential decay at 16\textsuperscript{th} and 22\textsuperscript{nd} epochs. For the slicing aided fine-tuning, patches are created by slicing the images and annotations and then Visdrone and xView training sets are augmented using these patches. Size of each patch is randomly selected to have a width and height in the range of 480 to 640 and 300 to 500 for Visdrone and xView datasets, respectively. Input images are resized to have a width of 800 to 1333 (by preserving the aspect ratio). During inference, NMS matching threshold $T_m$ is set as $0.5$.

The MS COCO \cite{lin2014microsoft} evaluation protocol has been adopted for evaluation, including overall and size-wise AP$_{50}$ scores. Specifically, AP$_{50}$ is computed at the single IoU threshold 0.5 over all categories and maximum number of detections is set as 500. In Table \ref{tab:visdrone-result} and \ref{tab:xview-result} conventional inference on original image, FI (Full Inference), is taken as the baseline. SF (Slicing Aided Fine-tuning) is the model fine-tuned on augmented dataset with patch sizes in the range of 480 to 640 and 300 to 500 in Tables \ref{tab:visdrone-result} and \ref{tab:xview-result}, respectively. SAHI (Slicing Aided Hyper Inference) refers to inference with patches of size $640\times640$ and $400\times400$ in Tables \ref{tab:visdrone-result} and \ref{tab:xview-result}, respectively. PO (Patch Overlap) means the there is 25\% overlap between patches during sliced inference.
As seen from Table \ref{tab:visdrone-result}, SAHI increases object detection AP by 6.8\%, 5.1\% and 5.3\%. The detection accuracy can be further increased with a SF, resulting in a cumulative increase of 12.7\%, 13.4\% and 14.5\% AP for FCOS, VFNet and TOOD detectors, respectively. Applying 25\% overlap between slices during inference, increases small/medium object AP and overall AP but slightly decreases large object AP. Increase is caused by the additional small object true positives predicted from slices and decrease is caused by the false positives predicted from slices that matching large ground truth boxes. Best small object detection AP is achieved by SF followed by SI, while best large object detection AP is achieved by SF followed by FI, confirming the contribution of FI for large object detection.
Results for xView dataset is presented in Table \ref{tab:xview-result}. Since xView targets are very small, regular training with original images yields poor detection performance and SF improves the results substantially. Integration of FI increases large object AP by up to 3.3\% but results in slightly decreased small/medium object AP, which is expected as some of the larger objects may not be detected from smaller slices. 25\% overlap between slices increase the detection AP by up to 1.7\%. xView contains highly imbalanced 60 target categories and despite being an older and, reportedly weaker detector, FCOS yields a very close performance compared to VFNet for this dataset. This observation confirms the effectiveness of focal loss \cite{lin2017focal} in FCOS, which is designed to handle category imbalance. TOOD also benefits from focal loss during training and yields the best detection result among 3 detector.
Error analysis results of TOOD detector on Visdrone and xView datasets are presented in Fig. \ref{fig:tood-visdrone-error} and \ref{fig:tood-xview-error}, respectively. Here C75, C50, Loc, Sim, Oth, BG, FN corresponds to results at IoU threshold of 0.75 and 0.50, results after ignoring localization errors, supercategory false positives, category confusions, all false positives, and all false negatives, respectively. As seen in Fig. \ref{fig:tood-visdrone-error}, there is minor room for improving super category false positives, category confusions and localization errors and major room for improving false positives and false negatives. Similarly, Fig. \ref{fig:tood-xview-error} shows that there is major room for improvement after fixing category confusions and false positives.



\vspace{-0.5cm}
\section{Conclusion}
\vspace{-0.2cm}
\label{sec:conc}

The proposed slicing aided hyper inference scheme can directly be integrated into any object detection inference pipeline and does not require pretraining. Experiments with FCOS, VFNet, and TOOD detectors on Visdrone and xView datasets show that it can result in up to 6.8\% AP increase. Moreover, applying slicing aided fine-tuning results in an additional 14.5\% AP increase for small objects and applying 25\% overlap between slices results in a further 2.9\% increase in AP. Training a network with higher resolution images through larger feature maps result in higher computation and memory requirements. The proposed approach increases the computational time linearly, while keeping memory requirements fixed. Computation and memory budgets can also be traded-off by adjusting the patch sizes, considering the target platform. In the future, instance segmentation models will be benchmarked utilizing the proposed slicing approach and different post-processing techniques will be evaluated.
 


\bibliographystyle{IEEEbib}
\small
\bibliography{strings,refs}

\end{document}